\documentclass[runningheads]{llncs}
\usepackage[T1]{fontenc}

\usepackage{graphicx}
\usepackage{booktabs}
\usepackage{makecell}

\usepackage[accsupp]{axessibility}
\usepackage{hyperref}
\usepackage{blindtext}
\usepackage{orcidlink}
\usepackage{color}

\begin{document}
\title{From Pixels to Words:\\Leveraging Explainability in Face Recognition through Interactive Natural Language Processing}
\titlerunning{From Pixels to Words}
%

\author{Ivan DeAndres-Tame\inst{1\orcidlink{0009-0008-5402-1153}}\and
Muhammad Faisal\inst{2}\and
Ruben Tolosana\inst{1\orcidlink{0000-0002-9393-3066}}\and
\\Rouqaiah Al-Refai\inst{2\orcidlink{0000-0001-8236-4593}}\and
Ruben Vera-Rodriguez\inst{1\orcidlink{0000-0002-6338-8511}}\and
Philipp Terh\"orst\inst{2}\orcidlink{0000-0001-8250-5712}}

\authorrunning{I.~DeAndres-Tame et al.}
%

\institute{Universidad Autonoma de Madrid, 28049 Madrid, Spain \and
Paderborn University, 33102 Paderborn, Germany}
\maketitle              
\begin{abstract}
Face Recognition (FR) has advanced significantly with the development of deep learning, achieving high accuracy in several applications. However, the lack of interpretability of these systems raises concerns about their accountability, fairness, and reliability. In the present study, we propose an interactive framework to enhance the explainability of FR models by combining model-agnostic Explainable Artificial Intelligence (XAI) and Natural Language Processing (NLP) techniques. The proposed framework is able to accurately answer various questions of the user through an interactive chatbot. In particular, the explanations generated by our proposed method are in the form of natural language text and visual representations, which for example can describe how different facial regions contribute to the similarity measure between two faces. This is achieved through the automatic analysis of the output's saliency heatmaps of the face images and a BERT question-answering model, providing users with an interface that facilitates a comprehensive understanding of the FR decisions. The proposed approach is interactive, allowing the users to ask questions to get more precise information based on the user's background knowledge. More importantly, in contrast to previous studies, our solution does not decrease the face recognition performance. We demonstrate the effectiveness of the method through different experiments, highlighting its potential to make FR systems more interpretable and user-friendly, especially in sensitive applications where decision-making transparency is crucial.

\end{abstract}
\section{Introduction}\label{sec:intro}

\begin{figure}[tb]
\centering
\includegraphics[width=\linewidth]{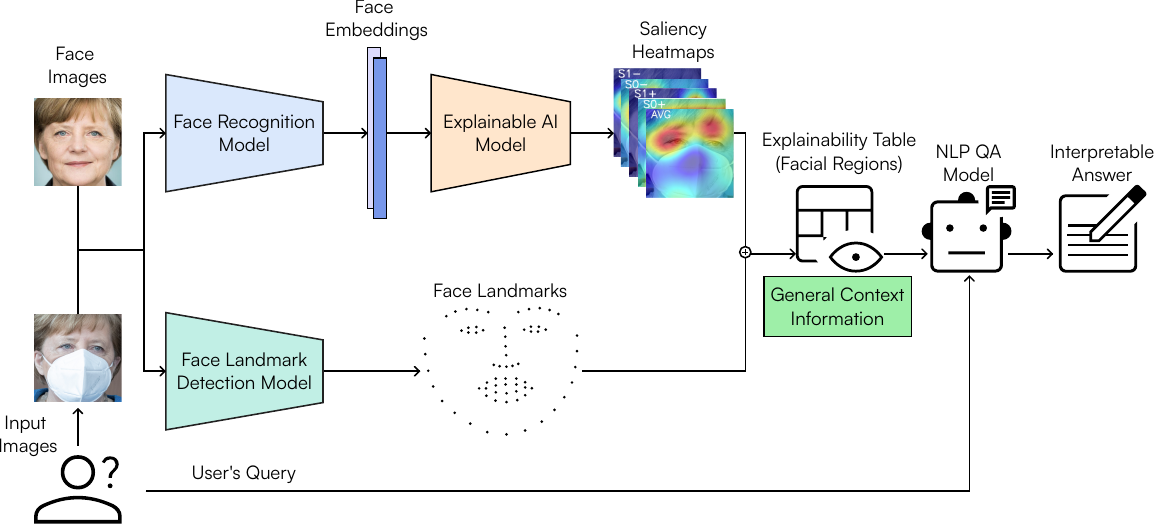}
\caption{Graphical representation of the proposed framework, focused on the combination of model-agnostic XAI and NLP techniques to leverage the explainability of the FR systems.}
\label{fig:method}
\end{figure}

Face biometrics is a popular field that analyzes multiple techniques mainly for identification and verification purposes~\cite{jain2006biometrics}. Several studies have shown its application to areas such as person recognition~\cite{wang2021deep,du2022elements}, synthetic face generation~\cite{melzi2023gandiffface,deandrestame2024second}, and security~\cite{li2022deep}, among others. In recent years, with the fast development of deep learning, significant advances have been made in Face Recognition (FR), surpassing previous benchmarks with impressive results~\cite{deng2019arcface,kim2022adaface,melzi2024frcsyn}. However, most of the state-of-the-art FR systems are based on Deep Neural Networks (DNN), which are often seen as black boxes that are hard to interpret, raising concerns about their fairness, accountability, and reliability~\cite{li2022interpretable}. Due to the lack of transparency of these DNN and the sensitivity of the biometric data, the explainability of FR models is becoming increasingly important, especially after the recent EU Artificial Intelligence Act\footnote{\url{https://artificialintelligenceact.eu/}}. Additionally, recent studies have proved the bias of the FR models, proposing new ways to mitigate it by interpreting the model's output~\cite{deandrestame2024how,terhoerst2022comprehensive}. 

The explainability of the models is especially important for real-life applications where choices can have a significant impact and a sensitive effect on society~\cite{adadi2018peeking}. As a result, leveraging explainability can bring non-technical users closer to understanding the operation of these FR models by making them easier to understand. Several Explainable Artificial Intelligence (XAI) techniques can be used for this purpose. However, as the complexity of state-of-the-art models cannot be covered only by these simple techniques, researchers have proposed model-agnostic techniques that explain the predictions or decisions made by a machine learning model, without needing to know how the model works internally \cite{lundberg2017unified,ribeiro2016why}. An example of such techniques applied to FR models is Grad-CAM~\cite{selvaraju2019grad}. Grad-CAM is a visual explanation technique used to represent which regions of an input image are crucial for the decision about a particular class. Grad-CAM produces a heatmap by weighting the feature maps with these gradients, highlighting the regions in the input image that most contribute to the final decision of a particular class. 

In particular, on the topic of FR, several approaches have been presented in the last few years. For example, in~\cite{lu2024towards,huber2024efficient} new explainable methods have been proposed to enhance the interpretability of FR systems through the most similar and dissimilar parts of the compared faces. Moreover, in~\cite{jiang2021explainable,neto2022explainable} they proposed different frameworks designed to enhance Explainable Face Recognition (XFR) while addressing the challenge of balancing explainability and recognition performance.
On the other hand, some studies have focused on analyzing the utility of single pixels in images to enhance explainability. For example, in~\cite{biagi2023explaining,terhoerst2023pixel} they introduced new methods to compute pixel-level face image quality, enhancing the explainability of face recognition models by detailing the contribution of each pixel to the overall utility for recognition. However, visual explanations alone are not user-friendly as they do not take care of the user's background knowledge and do not allow the user to ask questions to clarify the interpretation.

Recently, novel XAI techniques have been proposed in the literature based on Natural Language Processing (NLP). There are NLP models capable of providing a specific answer in natural language given a context and a question~\cite{allam2012question}. For example, BERT~\cite{devlin2018bert}, which stands for Bidirectional Encoder Representations from Transformers, uses a bidirectional network for language understanding. By leveraging a bidirectional transformer architecture, BERT is capable of learning information from both the left and right contexts, enabling a more comprehensive understanding of the relationships between words. In addition, recent approaches based on foundation models~\cite{achiam2023gpt,anil2023gemini} look promising, although they are still not adapted to specific applications such as FR, decreasing the performance in some scenarios~\cite{deandrestame2024how}. This fact motivates the present study, which aims to improve the explainability of the FR systems while keeping the same level of performance. 

In the present study, we propose an interactive framework to enhance the explainability of FR models by combining model-agnostic XAI and NLP techniques. Figure~\ref{fig:method} provides a graphical representation of the proposed framework. In particular, the explanations generated by our proposed framework are in the form of interactive natural language conversations. For this purpose, we use an explainability table, which describes how different regions of the face contribute to the similarity measure between two faces, together with some general context information about the FR system and decision, such as the similarity score, the confidence value, etc. This is achieved through the automatic analysis of the saliency heatmaps of the face images and a BERT Question-Answering (QA) model, providing users with an interface that facilitates a comprehensive understanding of the model's decisions. Our proposed explainability framework is characterized by:

\begin{itemize}
    \item \textbf{Interactive Explanations:} The proposed framework allows for a high human-machine interaction, enabling the user to ask questions to get more precise information depending on the user's background knowledge.
    \item \textbf{Model-Agnostic Framework:} The proposed framework can be easily applied to any FR model regardless of the specific architecture of the model, and thus maintaining the recognition performance.
    \item \textbf{Scalable:} The proposed framework already provides a variety of possible explanations based on the explainability table and general context information. In addition, the proposed framework can be easily extended with additional information that the user might be interested in without the need for retraining the system.
\end{itemize}

The remainder of the paper is organized as follows. Sec.~\ref{sec:method} describes the proposed method, including the details of the specific techniques used in each module. Sec.~\ref{sec:setup} explains all the details regarding our experimental setup, including the databases, metrics, and experimental protocol. Sec.~\ref{sec:results} provides the quantitative and qualitative results achieved by our proposed method. Finally, Sec.~\ref{sec:conclusions} draws the conclusions and points out future research lines.

\section{Proposed Method}\label{sec:method}

This section describes the proposed framework, which comprises three main modules: \textit{i)} the FR system and confidence estimation, described in Sec.~\ref{sub_fr}, \textit{ii)} the explainability method, described in Sec.~\ref{sub_explainability}, and finally \textit{iii)} the user-friendly QA interface based on NLP, described in Sec.~\ref{sub_NLP}.

\subsection{Face Recognition System and Confidence Estimation}\label{sub_fr}
To enable our framework to provide explainable conversations with a user, we first need to set up a FR system and a confidence estimation solution. Our framework can work with arbitrary FR systems and confidence estimation methods.

\begin{figure}[tb]
    \centering
    \includegraphics[width=\textwidth]{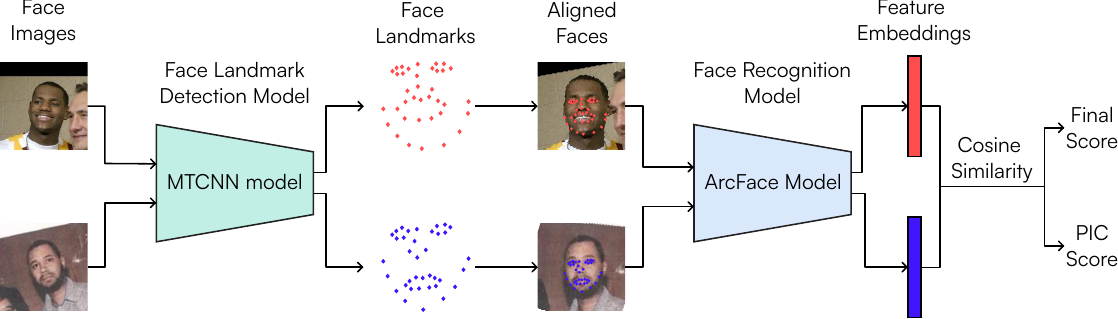}
    \caption{Graphical representation of the FR module. \cite{neto2023pic}.}
    \label{fig:fr_module}
\end{figure}

For our setup, to compare two face images, our proposed framework performs the face verification of the two face images, as described in Figure~\ref{fig:fr_module}. For this purpose, we first detect and align the faces in the images using the MTCNN model~\cite{zhang2016joint}. After extracting these landmarks, we align the original face images and forward them to the FR model in a frontal-view normalized pose. Regarding the FR model choice, we consider the ArcFace model~\cite{deng2019arcface} for its robustness and wide usage in different applications~\cite{wang2021deep,komkov2021advhat}. Finally, we compare the feature embeddings of the two face images using cosine similarity, to obtain a final score. If this score is above or below a decision threshold, the matching decision is classified as match (genuine) or non-match (impostor) \cite{wang2021deep,du2022elements}.

However, the final decision of the FR system is just a decision based on the cosine similarity, lacking the explainability of the decision. To increase the interpretability, we use the Probabilistic Interpretable Comparison (PIC) score~\cite{neto2023pic}. This metric is developed based on the concept of probabilistic comparison, whereby it evaluates the similarity score between multiple biometric samples, such as two face images, concerning the distribution of scores from the same identity and different identities. Its purpose is to determine the probability that the scores at hand are related to the same identity distribution rather than a different identity distribution. This metric provides a clear and meaningful probabilistic assessment of the matching confidence. Unlike previous methods that rely on arbitrary thresholds or scales, the PIC-Score can be directly understood as the determination of the likelihood that the matching decision is accurate. For example, a PIC-Score of 0.8 means that there is an 80\% chance that the two images belong to the same identity distribution. Note that when a pair is considered impostor (\textit{i.e.,} the face images are from different subjects), the probability for a correct decision is given by $1-$ PIC-Score, as indicated in~\cite{neto2023pic}.

\subsection{Explainability Method}\label{sub_explainability}
In addition to the PIC-Score described in the previous section, our proposed method increases the explainability of the FR system through: \textit{i)} a visual representation of the key facial regions for the decision, using saliency heatmaps, and \textit{ii)} an explainability table that includes values for the most and least important facial regions in the decision. For this purpose, we first generate saliency heatmaps~\cite{simonyan2013deep} from the face embeddings produced by the FR model. A saliency heatmap is a representation that highlights the most significant parts of an image or data, indicating where the attention should be focused. In the present study, we analyze five different saliency heatmaps based on the five different techniques presented in~\cite{mery2022black}. Examples of these techniques are shown in Figure~\ref{fig:xai_methods}. We explain next the key details for each of them:

\begin{figure}[tb]
    \centering
\includegraphics[width=\linewidth]{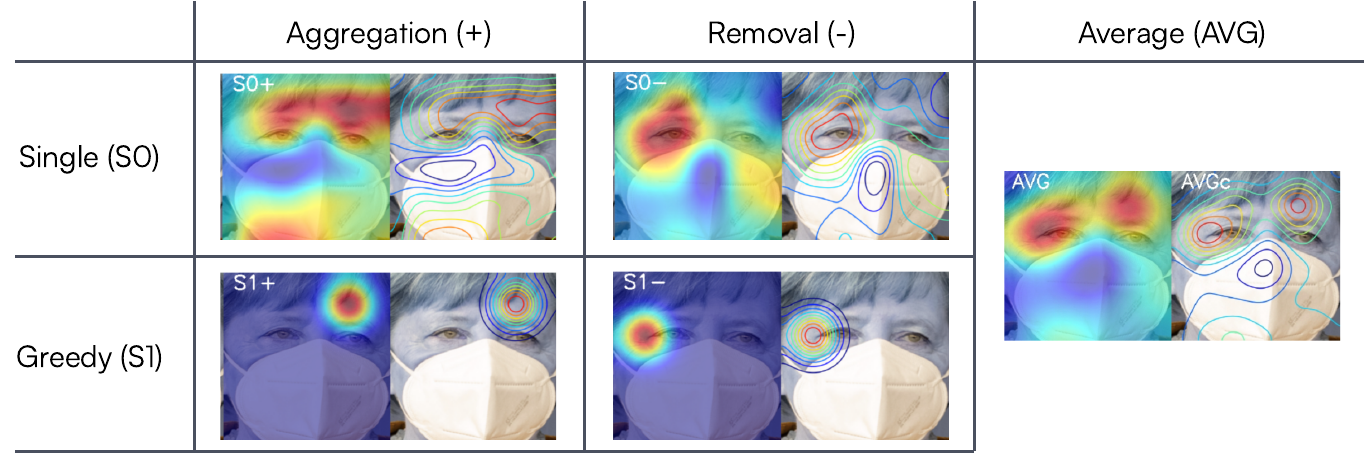}
    \caption{XAI methods presented in~\cite{mery2022black} to plot the saliency heatmap over a face image.}
    \label{fig:xai_methods}
\end{figure}

\begin{itemize}
\item  \textbf{Single Aggregation (S0+):} Finds the pixels that make the similarity score increase the most.
\item \textbf{Greedy Aggregation (S1+):} Adds the most important pixels one at a time, by repeating Single Aggregation several times. 
\item \textbf{Single Removal (S0-):} Finds the pixels that, when removed, make the similarity score drop the most.
\item \textbf{Greedy Removal (S1-):} Removes the most important pixels one by one, by repeating the Single Removal iteratively.
\item \textbf{Average Method (AVG):} This method calculates the average value across the above 4 methods.
\end{itemize} 

\begin{figure}[tb]
    \centering \includegraphics[width=\linewidth]{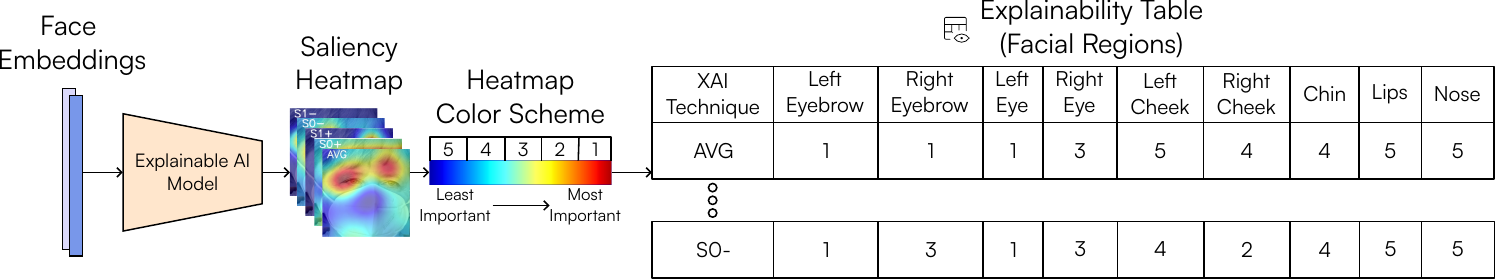}
    \caption{Graphical representation showing how the explainability table is created.}
    \label{fig:ExplainabilityTable}
\end{figure}

Finally, by aligning the saliency heatmaps with the facial landmarks, we are able to determine which facial regions are the most important ones in the final decision. The following facial regions are considered in the analysis: left and right eyebrows, left and right eyes, nose, left and right cheeks, chin, and lips. To better quantify the importance of each facial region, we transform the information of the saliency heatmaps into an explainability table. An example of this process can be seen in Fig.~\ref{fig:ExplainabilityTable}. To get a numerical value for each of the facial regions, we divide the output range of the saliency heatmaps into five intervals, assigning an importance score to each facial region based on the saliency heatmaps values. The scoring system spans from 1 to 5, where 1 represents the most important and 5 the least. The final values are included in the explainability table, providing a simple visual representation in terms of explainability.

\begin{figure}[tb]
    \centering \includegraphics[width=\linewidth]{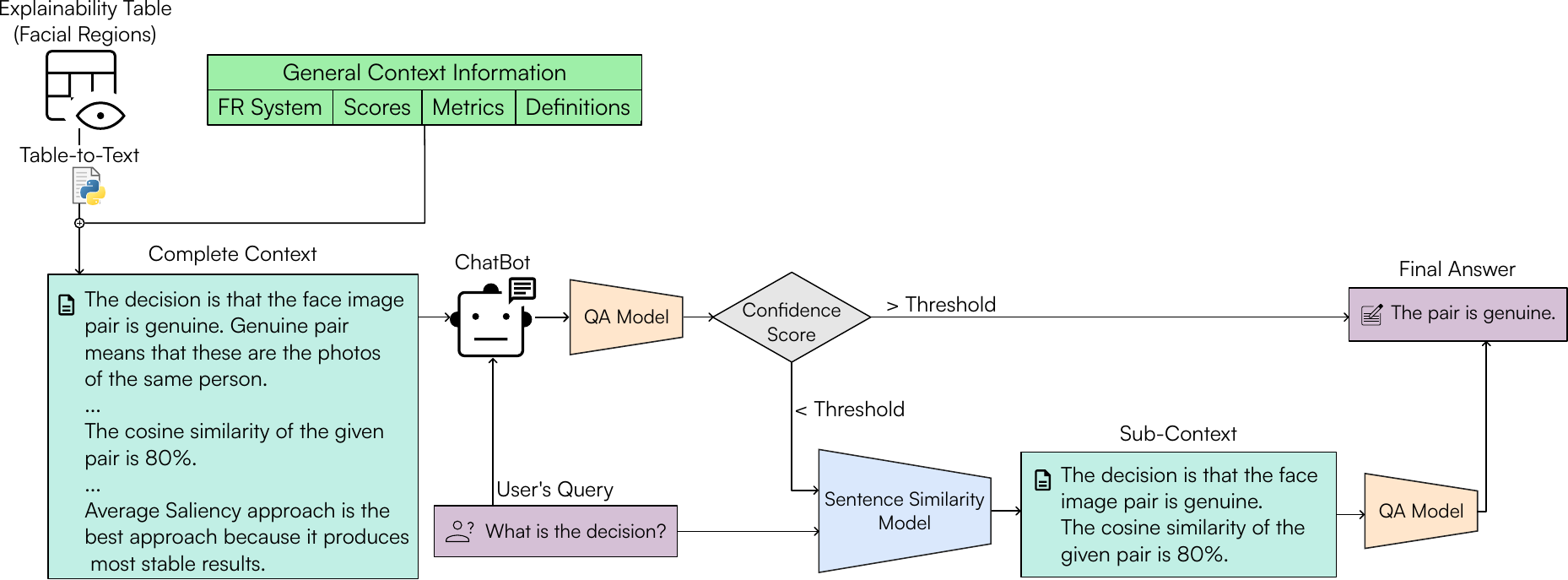}
    \caption{Graphical representation of the proposed QA interface based on the BERT NLP model~\cite{devlin2018bert}.}
    \label{fig:nlp-module}
\end{figure}

\subsection{Question-Answering Interface}\label{sub_NLP}
Finally, to provide a user-friendly and interpretable interface for all potential users, we propose integrating the FR system and confidence estimation (Sec.~\ref{sub_fr}) and explainability method (Sec.~\ref{sub_explainability}) into a QA interface. For this purpose, we use the BERT Question-Answering (BERT-QA) model\footnote{\href{https://huggingface.co/google-bert/bert-large-uncased-whole-word-masking-finetuned-squad}{https://huggingface.co/google-bert/bert-large-uncased-whole-word-masking-finetuned-squad}}~\cite{devlin2018bert}. This model takes two inputs: \textit{i)} the user's query, which is prompted by the user directly, and \textit{ii)} the context, which refers to the complete information of the matching process including the information needed to answer a specific query. Finally, the model returns a refined answer. Figure~\ref{fig:nlp-module} shows the complete pipeline for the proposed QA interface.

In our particular case, the context information introduced to the BERT-QA model is the explainability table with the importance of each facial region in the decision as well as the general context information. This includes details about the deployed FR system, the comparison score value, the decision, the corresponding confidence value, and general information about the explainability methods. Since the BERT-QA model cannot directly accept a table as input, we transform the explainability table and the general context information into plain text using a Python script. Therefore, the context contains the scores of all facial regions as well as all other information present in the table.

Regarding the functionality of the BERT-QA model, whenever the user asks a query, both the query and the context are sent to the BERT-QA model. The model processes the input and generates a response by understanding the context and extracting relevant information from it. The BERT-QA model outputs the answer along with a confidence score for the generated answer. If the confidence score of the answer is below a predefined threshold, it indicates uncertainty or ambiguity in the answer. This can happen due to a limitation of the BERT model, which does not always perform well with long input contexts. Therefore, if the confidence score is less than the threshold score, the answer is not accepted, and we ask the BERT-QA model the same query again but with a smaller context called sub-context. This sub-context is extracted from the original context and consists of a set of the most relevant sentences for the given query. These sentences are selected by calculating the similarity of the user query with each sentence in the complete context and choosing the ones with the highest similarity. For this purpose, we use the sentence embedding model MPNet-base~\cite{song2020mpnet}. This selected sub-context is then passed to the same BERT-QA model along with the user's query. The sub-context helps in obtaining a more focused and accurate response, allowing the BERT-QA model to generate a final answer and present it to the user.

\section{Experimental Setup}\label{sec:setup}

The experimental protocol and metrics considered in the present study have been designed to analyze quantitatively and qualitatively the performance of our end-to-end method, evaluating the three different modules.

\subsubsection{Face Recognition System.} For the experiments, we consider a pre-trained\footnote{\url{https://github.com/deepinsight/insightface/tree/master/recognition/arcface_torch}} FR model using the ArcFace loss \cite{deng2019arcface}.
The model is based on a ResNet-100 backbone trained on MS-Celeb-1MV3~\cite{guo2016ms}. To evaluate the performance of the FR model we use Labeled Faces in the Wild (LFW) database~\cite{huang2008labeled}. LFW is a very popular database in the FR field, containing over $13,000$ labeled face images from different real-life situations. These pictures include different lighting conditions, poses, facial emotions, and backgrounds, providing a benchmark to evaluate FR models in real-world conditions. Regarding the metrics of the FR system, we consider the Detection Error Tradeoff (DET) curve, the False Non-Match Rate (FNMR) for a False Match Rate (FMR) value of 0.01\% (high-security system), and the Equal Error Rate (EER).

\subsubsection{Explainability Method.} We analyze the variability of the scores generated by the different XAI techniques considered in the analysis (\textit{i.e.,} Single Removal, Greedy Removal, Single Aggregation, Greedy Aggregation, and the Average). In particular, the explainability analysis is carried out using the proposed explainability table, checking whether the majority of the XAI techniques focus on the same facial regions to make the final decision or not. The following metrics are calculated for each of the facial regions: \textit{i)} \textit{Mean}, defined as the average value of the different XAI techniques, and \textit{ii)} \textit{Ratio of 1s}, defined as the number of ``1'' values (\textit{i.e.}, most important facial region) observed in the different XAI techniques, divided by the total number of XAI techniques considered (5).  

\subsubsection{Question-Answering Interface.} To evaluate the performance of the proposed BERT-QA model for FR scenarios, we create a set of predefined queries to ask the model. These queries have different levels of complexity, going from simple ones like \textit{``What are the results of the comparison?''} to more in-depth ones that require the model to make use of most of the information included in the context, like \textit{``How did you come to this conclusion?''}. The evaluation of this module is performed qualitatively and quantitatively by checking the answers given by the model to each of the queries.


\section{Experimental Results}\label{sec:results}

\begin{figure}[t]
\centering
    \includegraphics[width=0.6\linewidth]{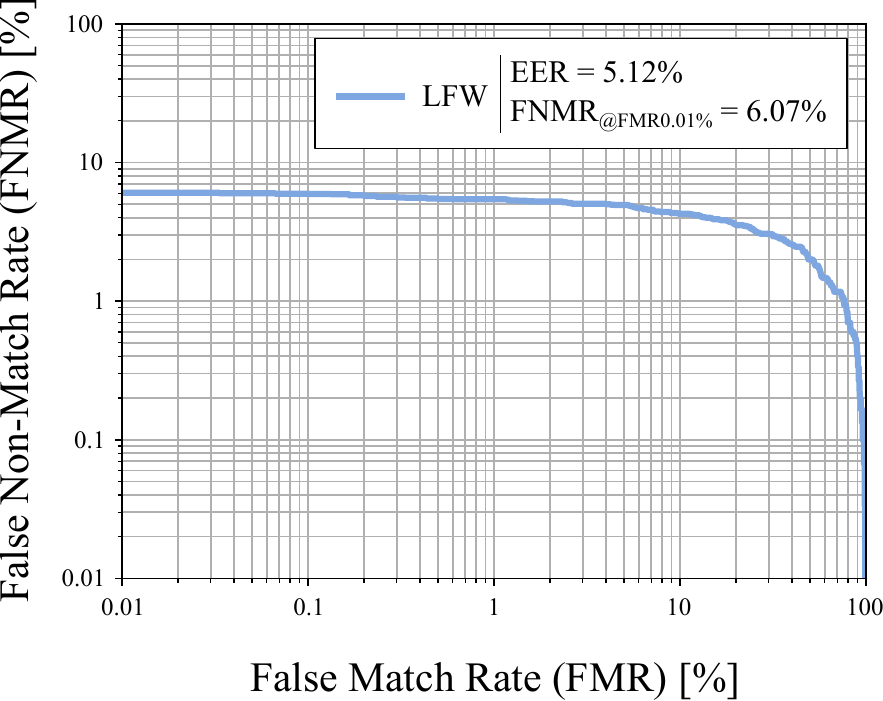}
    \caption{DET Curve of the verification performance of ArcFace~\cite{deng2019arcface} over LFW~\cite{huang2008labeled}.}
    \label{fig:det_curve}
\end{figure}

\subsection{Analysis of the Face Recognition System}\label{sec_experimentFR} 


First, we analyze the performance of the face verification system over LFW database. The DET curve for this database can be seen in Figure~\ref{fig:det_curve}. The EER is 5.12\%, while the FNMR value for a FMR fixed at $0.01\%$ is $6.07\%$. As can be seen, accurate results are achieved by the ArcFace model, proving to be reliable for high-security scenarios. Additionally, the model demonstrates to be balanced between false rejections and false acceptances, as indicated by the EER. This balance ensures that the system not only minimizes the risk of unauthorized access but also maintains user convenience by reducing the likelihood of false rejections.


\subsection{Analysis of the Explainability Method} 

Table~\ref{tab:table_xai} shows an example of two face images to compare, including the corresponding XAI techniques analyzed in the present study and the values inserted in the proposed explainability table. In this particular example, we intend to analyze the model's ability in challenging scenarios such as FR while wearing a face mask. As can be seen, each XAI technique provides valuable insights into the decision-making process by identifying the key facial regions. However, for some specific XAI techniques, such as Single Removal or Single Aggregation, we observe a higher variability in the explainability table depending on the facial region. To ensure a reliable approach, and select the optimal facial regions for the decision, we decide to consider the metrics Mean and Ratio of 1s values, which group the information from all the saliency heatmaps. The ratio of 1s simply describes what ratio of the XAI techniques has a score of 1 for the corresponding region. Following this approach, we can see that the eyebrows and eyes are selected as the most important facial regions for this particular example, as one of the face images is covered by a face mask.

\def\wimg{70px}

\begin{table}[t]
\caption{Example of the proposed explainability table, with the results obtained for the different saliency heatmaps and facial regions, including the final metrics: Mean and Ratio of 1s proposed to analyze the most discriminative facial regions.}\label{tab:table_xai}
\resizebox{\textwidth}{!}{\begin{tabular}{|l|c|c|c|c|c|c|c|}
\cline{2-6}

 \multicolumn{1}{c|}{} & \includegraphics[width=\wimg]{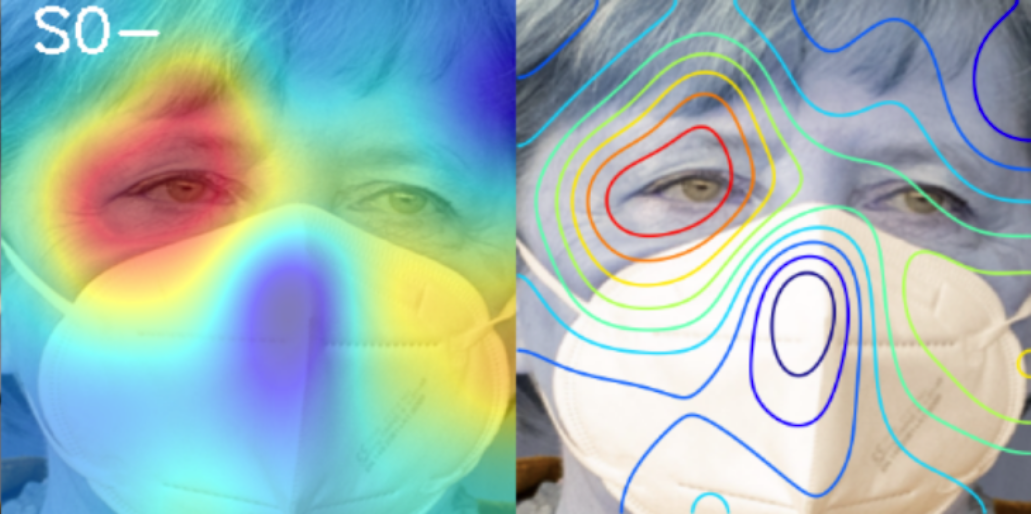} & \includegraphics[width=\wimg]{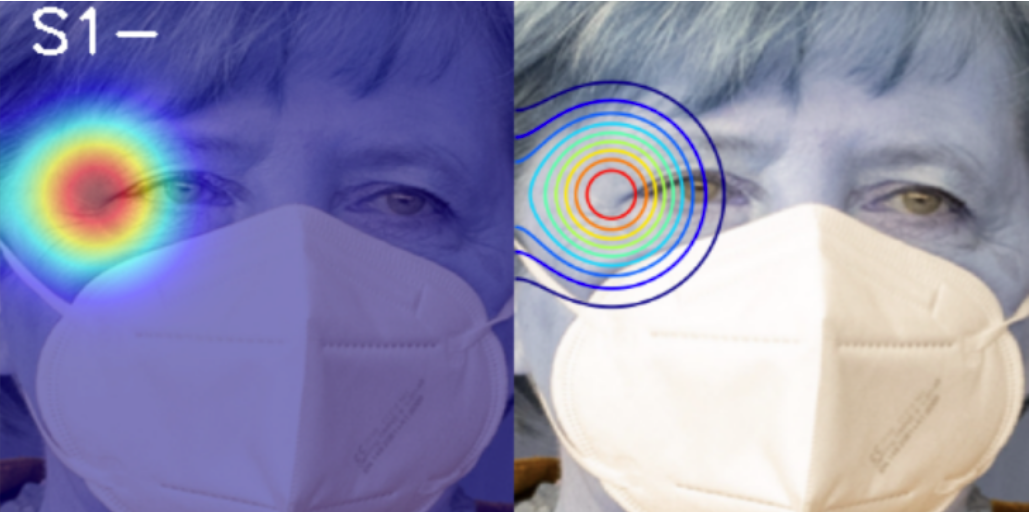} & \includegraphics[width=\wimg]{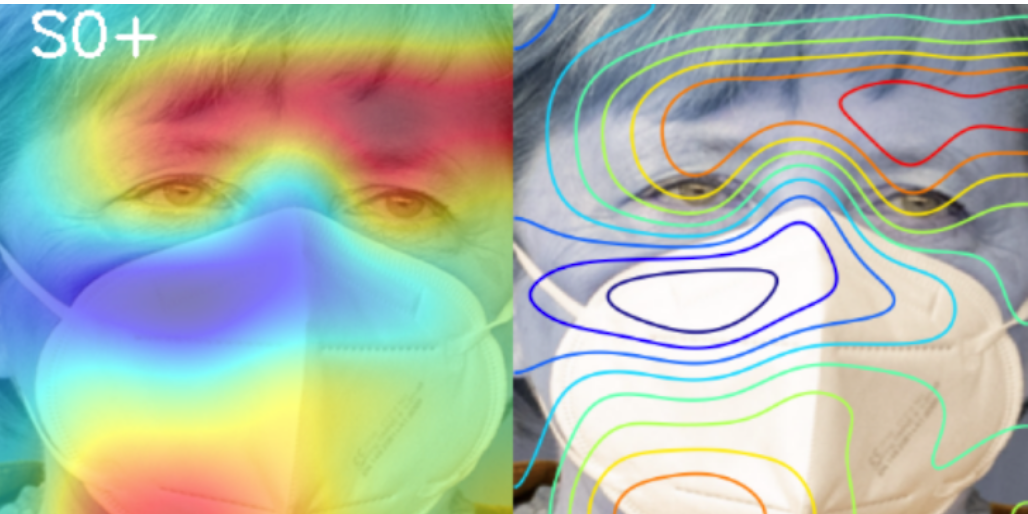} & \includegraphics[width=\wimg]{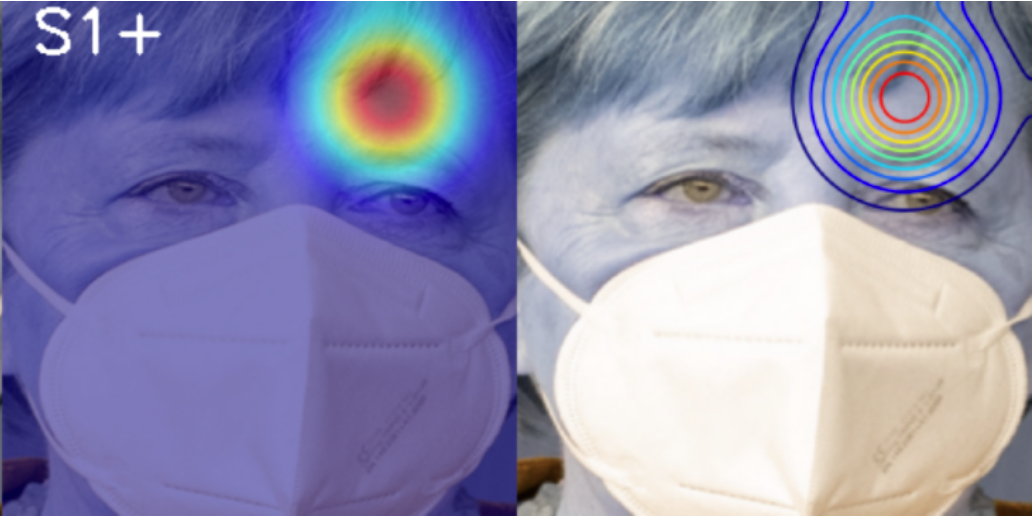} & \multicolumn{1}{c|}{\includegraphics[width=\wimg]{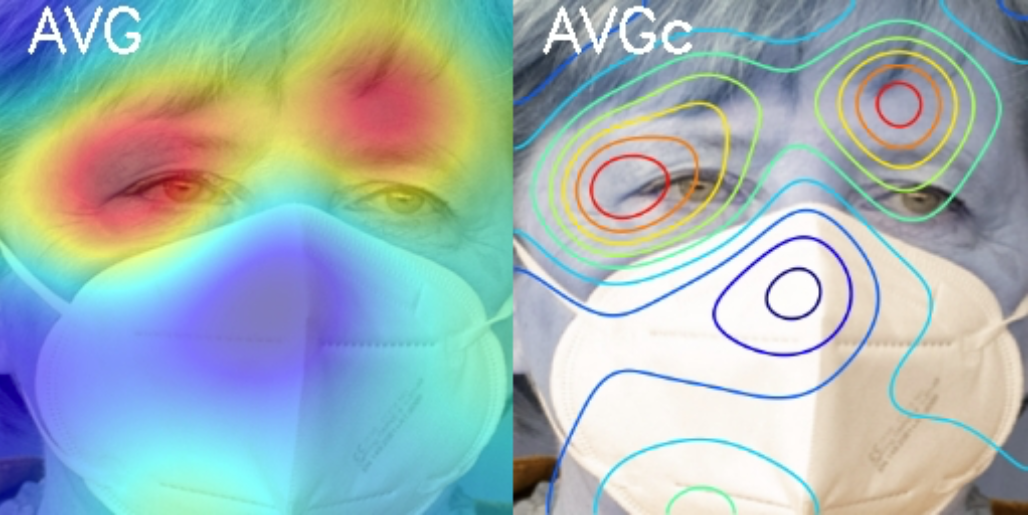}} & \multicolumn{2}{c}{} \\
 
\hline
\textbf{Facial Region} & \textbf{\begin{tabular}[c]{@{}c@{}}Single\\Removal \end{tabular}} & \textbf{\begin{tabular}[c]{@{}c@{}}Greedy\\Removal\end{tabular}} & \textbf{\begin{tabular}[c]{@{}c@{}}Single\\Aggregation\end{tabular}} & \textbf{\begin{tabular}[c]{@{}c@{}}Greedy \\ Aggregation\end{tabular}} & \textbf{Average} & \textbf{Mean $\downarrow$} & \textbf{\begin{tabular}[c]{@{}c@{}}Ratio\\ of 1s\end{tabular} $\uparrow$} \\ \hline
\textbf{Left Eyebrow} & \textbf{1} & \textbf{5} & \textbf{1} & \textbf{5} & \textbf{1} & \textbf{2.6} & \textbf{0.6} \\
 Right Eyebrow & 3 & 5 & 1 & 4 & 1 & 2.8 & 0.4 \\
 Left Eye & 1 & 5 & 2 & 5 & 1 & 2.8 & 0.4 \\
 Right Eye & 3 & 5 & 2 & 5 & 3 & 3.6 & 0.0 \\
 Left Cheek & 4 & 5 & 5 & 5 & 5 & 4.8 & 0.0 \\
 Right Cheek & 2 & 5 & 3 & 5 & 4 & 3.8 & 0.0 \\
 Chin & 4 & 5 & 1 & 5 & 4 & 3.8 & 0.2 \\
 Lips & 5 & 5 & 2 & 5 & 5 & 4.4 & 0.0 \\
 Nose & 5 & 5 & 5 & 5 & 5 & 5.0 & 0.0 \\ \hline
\end{tabular}
}
\end{table}


\begin{figure}[!]
    \centering    \includegraphics[width=\linewidth]{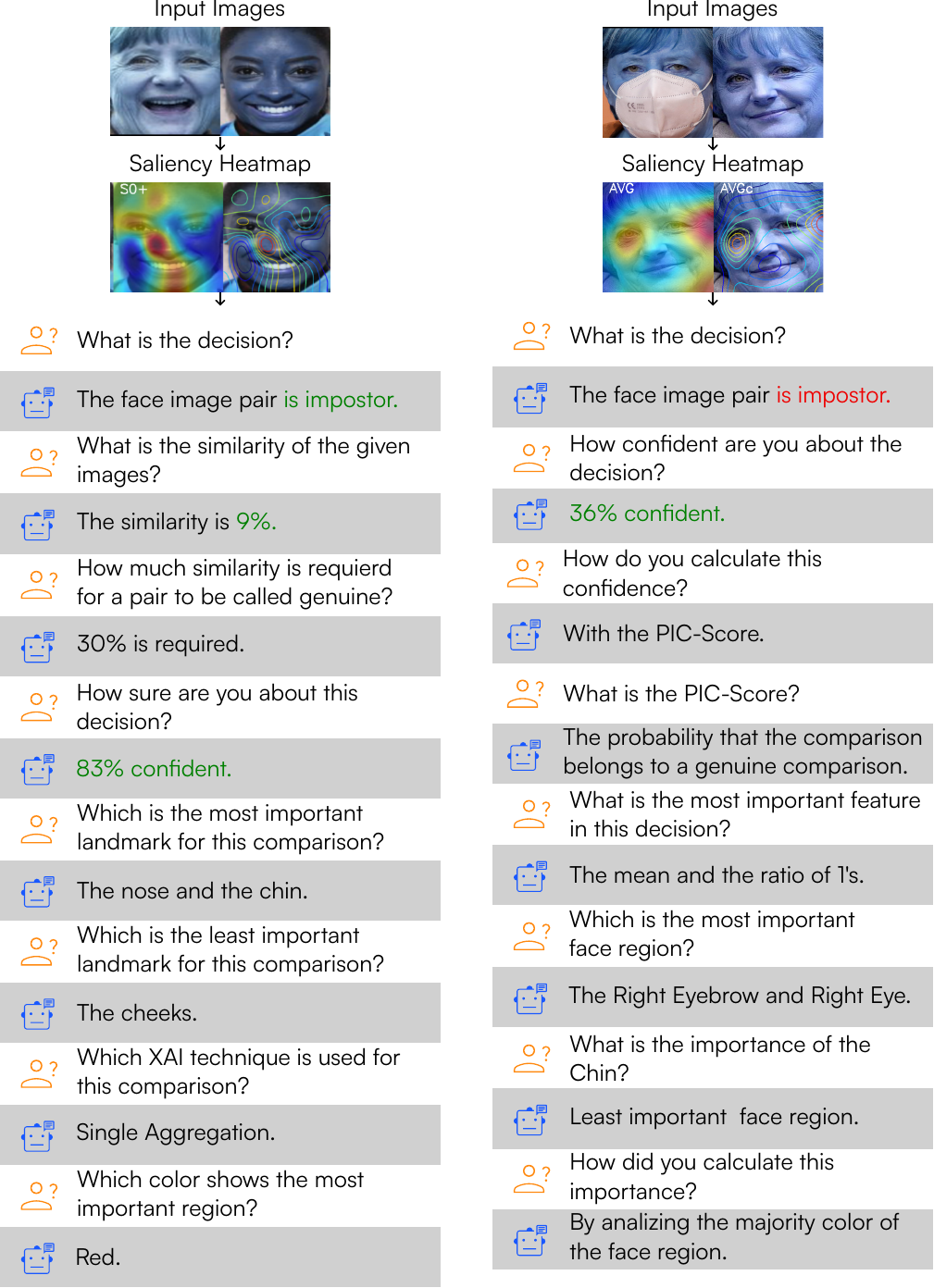}
    \caption{Graphical examples of the user interaction with the chatbot (BERT-QA model), introducing wrong predicted pairs (impostor labeled as genuine and vice versa). By checking the confidence value of the decision or inspecting the correctness of the reasoning, the user can easily detect that something is wrong with the decision, allowing for a manual review through the saliency heatmaps.}
    \label{fig:chat_comb}
\end{figure}


\subsection{Analysis of the Question-Answering Interface}\label{sec_experimentsQA}

\subsubsection{Qualitative Analysis.}

 Figure~\ref{fig:chat_comb} provides graphical examples of the user interaction with the chatbot (BERT-QA model), introducing two impostor comparisons as input. As can be seen on the left chat, our method correctly predicts the result of the comparison with high confidence and the chatbot is able to provide valuable insights and explanations on how the decision is made. However, on the right chat, we provide an examples in which the prediction on the pair images is wrong. Users can prompt various questions about the facial recognition results, decision-making process, confidence levels, metrics, landmarks, etc., and through reviewing the chatbot answers (\textit{e.g.,} the confidence score) detect that the prediction is wrong and manually review it through the saliency heatmaps, for example. The proposed method provides a user-friendly interface for interpreting FR results, ensuring transparency and enhancing user understanding of the complex FR system. This interactive approach bridges the gap between technical outputs and user comprehension, fostering trust and usability, especially for users with no experience in the topic.

\begin{table}[t]
\centering
\renewcommand{\arraystretch}{1.1}
\caption{Quantitative analysis of the QA module. For each question, 16 variants of these questions are generated and the corresponding answers were evaluated for correctness. For some questions, a technically correct answer was provided that however did not answer the meant questions. For these scenarios, we additionally analyzed the correctness with a follow-up question making the question more precise.}
\label{tab:Quantitative analysis}
\resizebox{0.9\linewidth}{!}{\begin{tabular}{lc}
\Xhline{2\arrayrulewidth}
\textbf{Question Prompt}                                    & \textbf{Correctness Rate [\%]} \\ 
\hline
What is the decision?                   & \phantom{1}68.75\%                  \\
What is the decision? (with follow-up question)              & 100.00\%                    \\
How did you come to this decision?       & \phantom{1}75.00\%                     \\
How did you come to this decision? (with follow-up question) & 100.00\%                    \\
How sure are you about the decision?                    & 100.00\%                    \\
What is the most important feature of this decision?    & \phantom{1}87.00\%                     \\
What is the least important feature of this decision?   & 100.00\%                    \\
What is explainable AI?                                 & 100.00\%                    \\
What are these output images?                               & 100.00\%                    \\ 
\Xhline{2\arrayrulewidth}
\end{tabular}}
\end{table}

\subsubsection{Quantitative Analysis.}

For completeness, we quantitatively analyze the correctness and stability of the answers provided by the chatbot. This analysis is shown in Table~\ref{tab:Quantitative analysis}. For the experiment, we generated 16 variants of each of the nine questions shown in the table. Then, each question variant is tested on a random image pair resulting in 144 conversations that were manually checked on random image pairs. This allows us to analyze how stable the QA module is regarding different sentence structures and provides an indicator of the correctness of its answers. The questions asked vary from questions about the decision (``What is the decision?''), questions about the decision confidence (``How sure are you about this decision?''), technical questions (``What is explainable AI?''), up to questions about the reason for the decision (``What is the most important feature of this decision?''). These questions were chosen to enable the user to make an informed decision towards trusting or distrusting the system's decision based on its reasoning.

The results in Table~\ref{tab:Quantitative analysis} show that technical questions and questions about decision confidence provided perfect answers (100\% correctness). For questions about the reasoning of the decision-making process, the correctness was still quite high (with 87\% and 100\% correctness). For two question scenarios, the model provided technically correct answers but did not provide the expected answer. For these cases, we additionally analyzed the performance with a follow-up question specifying the meaning of the questions. For instance, on the question variants for ``What is the decision?'', in 31.25\% of the cases the answer was a technical definition of matching decision instead of getting the matching decision of the given face image pair. With a follow-up question to specify the meaning of the question, \textit{e.g.,} ``What is the decision \textit{of the given image pair}?'', the correctness went to 100\%. This shows a big advantage of having a chatbot, as the user can interactively specify the meaning of a question and the chatbot can react to it by providing a more meaningful answer.

\section{Conclusions}\label{sec:conclusions}
The present study has focused on the critical challenge of explaining the decisions made by FR systems, as current FR systems are often seen as black boxes hard to interpret. In particular, our proposed framework enhances the explainability of the deployed FR systems by combining model-agnostic XAI and NLP techniques. As a result, the explanations generated by our proposed framework are in the form of natural language text and visual representations. This is achieved through the automatic analysis of the output's saliency heatmaps of the face images and a BERT-QA model, providing users with an interface that facilitates a comprehensive understanding of the FR decisions. 

To summarize, our proposed explainability framework is characterized by: \textit{i)} interactive explanations, enabling users to ask questions to get more precise information depending on the user's background knowledge; \textit{ii)} model-agnostic scheme, which can be easily applied to any FR model regardless of the specific architecture of the model; \textit{iii)} high recognition performance, unlike recent popular foundation models such as ChatGPT whose performance considerably degrades in some application scenarios~\cite{deandrestame2024how,ahmad2024chatgpt}, as our solution does not alter the deployed FR system; and \textit{iv)} scalability, as the proposed framework can be easily extended with additional information that the user might be interested in without the need for retraining the system.

{\small \textbf{Acknowledgments.} 
This study has received funding from INTER-ACTION \\(PID2021-126521OB-I00 MICINN/FEDER), Cátedra ENIA UAM-VERIDAS en IA Responsable (NextGenerationEU PRTR TSI-100927-2023-2), R\&D Agreement\\DGGC/UAM/FUAM for Biometrics and Cybersecurity, and PowerAI+ (SI4/PJI/2024-00062, funded by Comunidad de Madrid through the grant agreement for the promotion of research and technology transfer at UAM).}

%
%
%
\bibliographystyle{splncs04}
\bibliography{main}
\end{document}